\providecommand{\tightlist}{%
  \setlength{\itemsep}{0pt}\setlength{\parskip}{0pt}}
\documentclass{article}
\usepackage{arxiv} % If you are using the arxiv.sty style
\usepackage[T1]{fontenc}
\usepackage[utf8]{inputenc}
\usepackage{amsmath,amssymb}
\usepackage{graphicx}
\usepackage{hyperref}
\usepackage{libertine} % Linux Libertine text font (pdflatex compatible)
\usepackage[libertine]{newtxmath} % Math font matching Libertine
\usepackage{xcolor}
\usepackage{framed}
\usepackage{booktabs} % For professional tables
\definecolor{shadecolor}{RGB}{248,248,248}

\title{Statistical Properties of the King Wen Sequence:\\An Anti-Habituation Structure That Does Not Improve Neural Network Training}
\author{Augustin Chan\\
\texttt{aug@iterative.day}\\
Independent Researcher}
\date{\today}

\begin{document}
\maketitle

\begin{abstract}
The King Wen sequence of the I-Ching (c.\ 1000 BC) orders 64 hexagrams---states of a six-dimensional binary space---in a pattern that has puzzled scholars for three millennia. We present a rigorous statistical characterization of this ordering using Monte Carlo permutation analysis against 100,000 random baselines. We find that the sequence has four statistically significant properties: higher-than-random transition distance (98.2nd percentile), negative lag-1 autocorrelation ($p=0.037$), yang-balanced groups of four ($p=0.002$), and asymmetric within-pair vs.\ between-pair distances (99.2nd percentile). These properties superficially resemble principles from curriculum learning and curiosity-driven exploration, motivating the hypothesis that they might benefit neural network training. We test this hypothesis through three experiments: learning rate schedule modulation, curriculum ordering, and seed sensitivity analysis, conducted across two hardware platforms (NVIDIA RTX 2060 with PyTorch and Apple Silicon with MLX). The results are uniformly negative. King Wen LR modulation degrades performance at all tested amplitudes. As curriculum ordering, King Wen is the worst non-sequential ordering on one platform and within noise on the other. A 30-seed sweep confirms that only King Wen's degradation exceeds natural seed variance. We explain why: the sequence's high variance---the very property that makes it statistically distinctive---destabilizes gradient-based optimization. Anti-habituation in a fixed combinatorial sequence is not the same as effective training dynamics.
\end{abstract}

\textbf{Keywords:} King Wen sequence, I-Ching, curriculum learning, information-theoretic surprise, negative results, learning rate scheduling, reproducibility

%% ============================================================
\section{Introduction}\label{sec:introduction}
%% ============================================================

The King Wen sequence, traditionally dated to approximately 1000 BC, orders the 64 hexagrams of the I-Ching in a pattern that has long defied mathematical explanation. Unlike the Shao Yong ordering (c.\ 1050 AD), which follows a straightforward binary enumeration, the King Wen sequence exhibits no obvious numerical or algebraic rule. The mathematical significance of its structure was first recognized by Leibniz \cite{leibniz1703binary}, who discovered parallels between the binary number system and the I-Ching's hexagrams.

In recent years, the I-Ching has found applications in computational intelligence, from evolutionary algorithms \cite{chen2016iching} to neural architecture search \cite{zhang2021asnas}. More broadly, structured knowledge systems outside conventional machine learning have shown measurable impacts on AI system performance \cite{choi2023confirm}.

We begin by establishing that the King Wen sequence has genuine statistical structure. Using Monte Carlo permutation analysis against 100,000 random baselines, we confirm four properties that distinguish it from chance orderings: high transition distance, negative autocorrelation, local yang balance, and pair-level asymmetry. These properties superficially resemble principles from curriculum learning \cite{bengio2009curriculum,graves2017automated} and curiosity-driven exploration \cite{schmidhuber2006developmental}. A natural hypothesis follows: does the sequence's anti-habituation profile translate to measurable advantages in neural network training?

We test this hypothesis rigorously and find that it does not hold. Across three experiments---learning rate schedule modulation, curriculum ordering, and seed sensitivity analysis---the King Wen sequence either degrades training performance or produces results indistinguishable from noise. We explain the mechanism: the sequence's high variance, which generates its distinctive statistical signature, is counterproductive for gradient-based optimization.

We make four contributions:

\begin{enumerate}
\tightlist
\item An enhanced statistical characterization of the King Wen sequence via 100,000-permutation Monte Carlo analysis, confirming four statistically significant combinatorial properties.
\item Learning rate schedule modulation experiments showing that the sequence's variance profile degrades training across all tested amplitudes.
\item Curriculum ordering experiments on two hardware platforms, isolating a \texttt{torch.compile} confound from genuine learning dynamics.
\item A mechanistic explanation of why anti-habituation statistical properties do not entail training benefit, grounded in the gap between combinatorial structure and optimization dynamics.
\end{enumerate}

%% ============================================================
\section{Related Work}\label{sec:related}
%% ============================================================

\subsection{Curriculum Learning}

Bengio et al.\ \cite{bengio2009curriculum} formalized the intuition that training machine learning models in a meaningful order---from simple to complex examples---improves convergence and generalization. Graves et al.\ \cite{graves2017automated} extended this to automated curriculum learning, using multi-armed bandit algorithms to select training tasks. Wang et al.\ \cite{wang2021curriculum} survey the field, distinguishing difficulty measurers from training schedulers. Recent work has shown that curriculum ordering interacts strongly with learning rate schedules \cite{luo2025lrdecay} and that simple difficulty metrics (compression ratio, lexical diversity) are effective at scale \cite{zhang2025curriculum}.

\subsection{I-Ching in Computational Intelligence}

Chen et al.\ \cite{chen2016iching} proposed the I-Ching Divination Evolutionary Algorithm (IDEA), using hexagram transformation operations as evolutionary operators. Zhang et al.\ \cite{zhang2021asnas} extended this to neural architecture search with AS-NAS. These works use the I-Ching's transformation mechanics as algorithmic primitives. Our investigation is different: we test whether the King Wen sequence ordering itself---not the hexagram transformation rules---has optimization utility.

\subsection{Information-Theoretic Surprise and Optimal Learning}

Schmidhuber \cite{schmidhuber2006developmental} proposed that artificial curiosity should drive agents toward states of intermediate surprise. Itti and Baldi \cite{itti2009bayesian} formalized Bayesian surprise as KL divergence between prior and posterior beliefs. Nielsen \cite{Nielsen_2020} provides the information-geometric framework connecting these ideas. We test whether the King Wen sequence's surprise profile---which exhibits properties consistent with these frameworks---produces practical benefits when applied to neural network training.

\subsection{Value of Negative Results}

Negative results are essential for scientific progress yet remain underreported in machine learning. When a hypothesis has surface plausibility---as the King Wen sequence does, given its genuine statistical properties---rigorous negative results prevent others from pursuing similar dead ends and clarify the boundary between theoretical elegance and practical utility. Our contribution follows this tradition: we test a natural hypothesis with proper controls and report that it fails, with mechanistic explanation.

%% ============================================================
\section{The King Wen Sequence: Statistical Characterization}\label{sec:stats}
%% ============================================================

\subsection{Hexagram Space and Notation}

Each hexagram is a six-bit binary vector representing six lines (yin${}=0$ or yang${}=1$), arranged from bottom to top. The 64 hexagrams span the complete space $\{0,1\}^6$. The King Wen sequence assigns each hexagram a position from 1 to 64 in a traditional ordering. We also consider the Shao Yong ordering (c.\ 1050 AD), which arranges hexagrams by binary value using a reversed-trigram convention, and the natural binary ordering (hexagrams 0--63 by integer value).

\subsection{Transition Metrics}

For consecutive hexagrams $H_i$ and $H_{i+1}$ in a sequence, we measure:

\begin{itemize}
\tightlist
\item \textbf{Hamming distance} $d_H(H_i, H_{i+1})$: the number of differing bit positions (range 0--6)
\item \textbf{Trigram relationship distance}: distance between upper and lower trigram pairs
\item \textbf{Nuclear hexagram distance}: Hamming distance between the inner four lines (positions 2--5)
\end{itemize}

We define information-theoretic surprise as:
\begin{equation}
S(H_i, H_{i+1}) = -\log P(H_{i+1} \mid H_i)
\end{equation}
where conditional probability is modeled via pattern similarity incorporating traditional line position weights $[0.03, 0.07, 0.15, 0.20, 0.25, 0.30]$ from bottom to top, with yang-to-yin transitions weighted at $0.7\times$ the reverse. Total similarity combines external (line-level) and internal (nuclear hexagram) components with $\lambda = 0.4$ weighting the nuclear component.

\begin{figure}[ht]
\centering
\includegraphics[width=\linewidth]{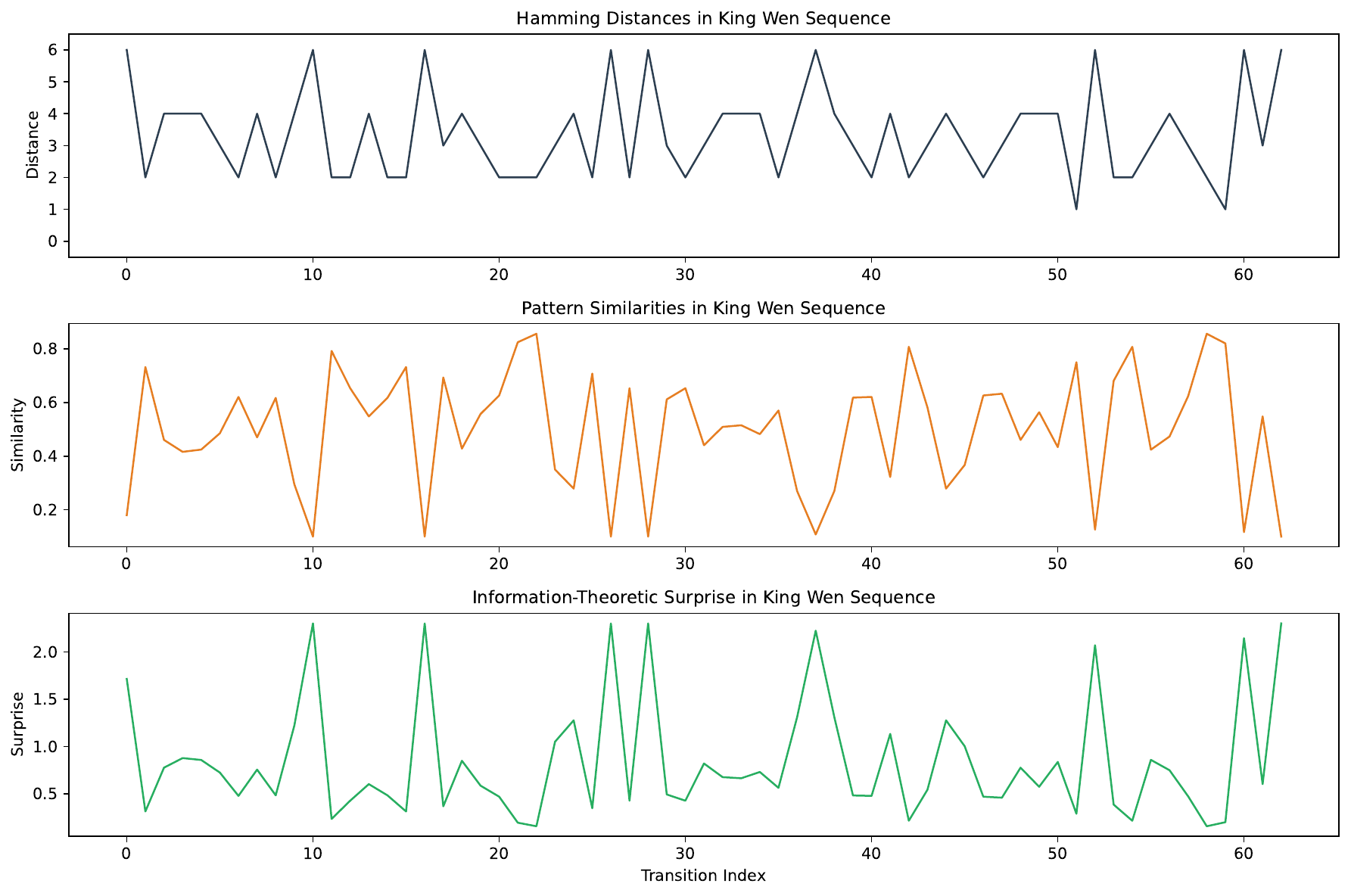}
\caption{Transition metrics across the 63 consecutive hexagram transitions of the King Wen sequence. \textbf{Top:} Hamming distance (number of differing lines, range 0--6). \textbf{Middle:} pattern similarity incorporating traditional line-position weights. \textbf{Bottom:} information-theoretic surprise $S(H_i, H_{i+1}) = -\log P(H_{i+1} \mid H_i)$. The surprise profile alternates between high and low values, consistent with the sequence's negative lag-1 autocorrelation.}
\label{fig:metrics}
\end{figure}

Figure~\ref{fig:metrics} shows these three transition metrics---Hamming distance, pattern similarity, and information-theoretic surprise---across the King Wen sequence's 63 consecutive transitions.

\subsection{Monte Carlo Permutation Analysis}

We compare the King Wen sequence against 100,000 random permutations of the same 64 hexagrams. Four properties are statistically significant:

\textbf{1.\ Higher-than-random transition distance (98.2nd percentile).} Mean Hamming distance between consecutive hexagrams: 3.35 (random mean: 3.05, $\sigma = 0.15$). The sequence deliberately maximizes change between consecutive states.

\textbf{2.\ Negative lag-1 autocorrelation (3.7th percentile, $p = 0.037$).} Autocorrelation of Hamming distances at lag 1: $-0.251$ (random mean: $-0.032$, $\sigma = 0.124$). A large transition is systematically followed by a small one and vice versa. This alternation between dramatic and subtle change is actively constructed, not a property of random permutations.

\textbf{3.\ Yang-balanced groups of four (99.8th percentile, $p = 0.002$).} Seven out of 16 groups of four consecutive hexagrams have exactly 12 yang lines (perfect yin-yang balance). Random expectation: 2.6 ($\sigma = 1.5$). The sequence maintains energetic equilibrium at the 4-hexagram scale.

\textbf{4.\ Within-pair vs.\ between-pair asymmetry (99.2nd percentile).} Within-pair mean Hamming distance: 3.56. Between-pair: 2.94. Asymmetry: 0.63. Paired hexagrams are maximally different from each other (complement/inverse), while transitions between pairs are smoother---creating tension within pairs and resolution between them.

\subsection{Comparison with Systematic Orderings}

\begin{figure}[ht]
\centering
\includegraphics[width=\linewidth]{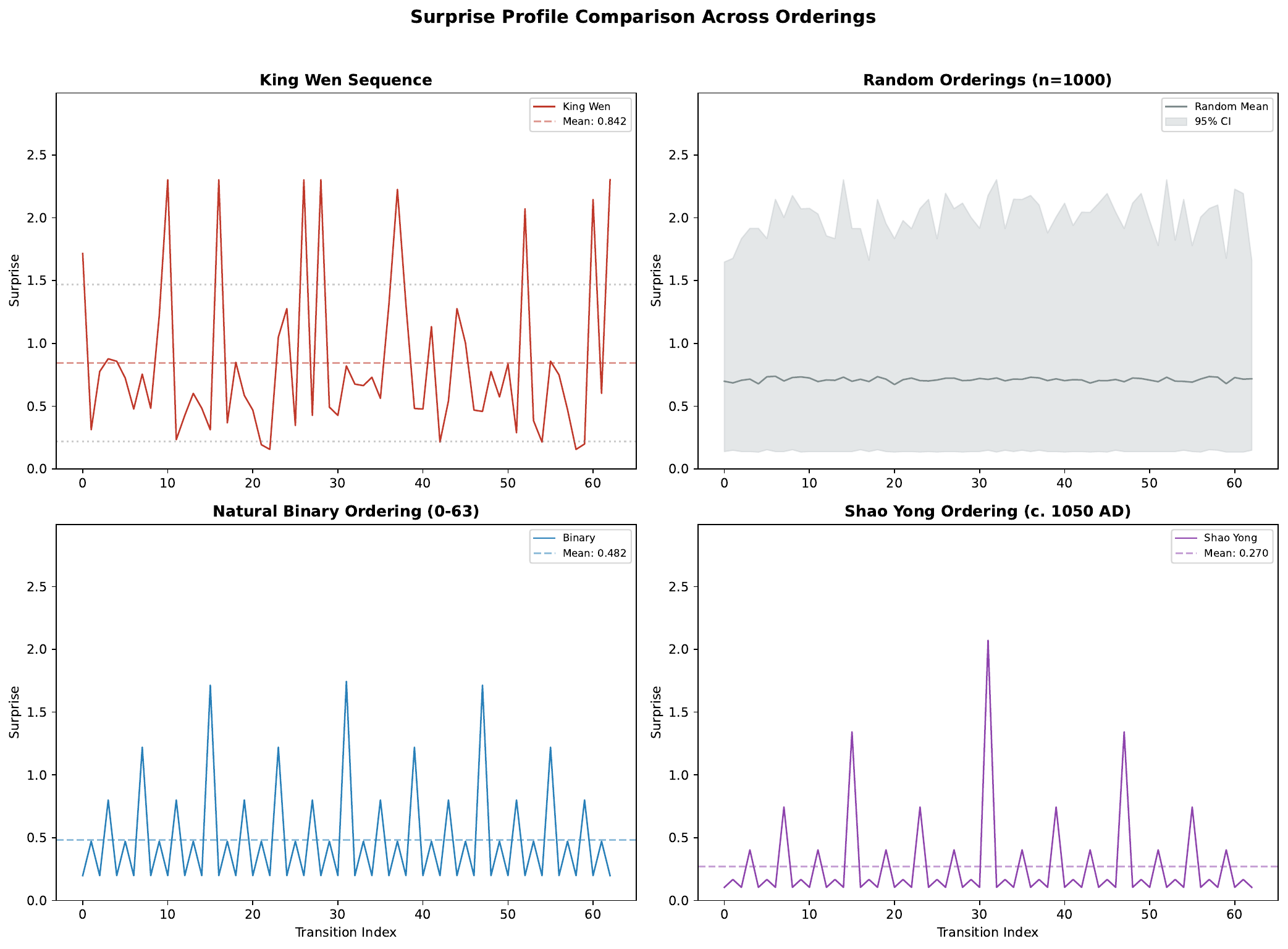}
\caption{Information-theoretic surprise across all 63 transitions for four orderings. \textbf{Top-left:} King Wen, with mean (dashed) and $\pm 1\sigma$ band, showing high-variance, full-spectrum surprise. \textbf{Top-right:} mean and 95\% confidence interval over 1{,}000 random permutations. \textbf{Bottom-left:} natural binary ordering. \textbf{Bottom-right:} Shao Yong ordering. King Wen resembles the random ensemble in mean surprise but exhibits higher variance, whereas binary and Shao Yong are smoother and strongly autocorrelated.}
\label{fig:comparison}
\end{figure}

We compare the surprise distributions of four orderings across all 63 consecutive transitions (Table~\ref{tab:stats}, Figure~\ref{fig:comparison}).

\begin{table}[h]
\centering
\begin{tabular}{lcccc}
\toprule
\textbf{Ordering} & \textbf{Mean} & \textbf{Std} & \textbf{Variance} & \textbf{Range} \\
\midrule
King Wen    & 0.842 & 0.624 & 0.390 & 0.16--2.30 \\
Random (mean of 1000) & 0.711 & 0.453 & 0.202 & 0.11--2.30 \\
Binary      & 0.482 & 0.403 & 0.162 & 0.20--1.75 \\
Shao Yong   & 0.270 & 0.348 & 0.121 & 0.11--2.07 \\
\bottomrule
\end{tabular}
\caption{Summary statistics for surprise distributions across orderings.}
\label{tab:stats}
\end{table}

\textbf{Distribution shape (Kolmogorov-Smirnov test):} King Wen is not significantly different from random ($D = 0.11$, $p = 0.44$), but is highly significantly different from binary ($D = 0.46$, $p < 0.001$) and Shao Yong ($D = 0.73$, $p < 0.001$).

\textbf{Variance (Levene's test):} King Wen has significantly higher variance than random ($p = 0.009$) and Shao Yong ($p < 0.001$). The difference versus binary is marginal ($p = 0.053$).

\textbf{Autocorrelation (Ljung-Box, lag 5):} King Wen shows no significant autocorrelation ($p = 0.45$), while binary ($p < 0.001$) and Shao Yong ($p = 0.002$) are highly autocorrelated. Only 3.9\% of random permutations show significant autocorrelation.

\subsection{Summary}

The King Wen sequence is neither random nor algebraically systematic. It occupies a distinctive position: random-like mean surprise, significantly higher variance, negative lag-1 autocorrelation, and local yang balance. These properties are real and confirmed against rigorous baselines. Whether they are useful for optimization is a separate, empirical question---addressed in the next section.

%% ============================================================
\section{Experimental Evaluation}\label{sec:experiments}
%% ============================================================

We test the hypothesis that the King Wen sequence's anti-habituation properties improve neural network training. We conduct three experiments using the autoresearch framework---Karpathy's lightweight experiment runner, together with an MLX port for the Apple Silicon experiments (Appendix~\ref{app:code})---which runs fixed-duration (5-minute) training experiments on a small GPT language model and evaluates using validation bits per byte (val\_bpb; lower is better).

\subsection{Experimental Setup}

\textbf{Model.} GPT with rotary position embeddings (RoPE), sliding window attention, and a hybrid optimizer (Muon for matrix parameters, AdamW for embeddings and scalars). Default configuration: DEPTH${}=4$ (${\sim}11.5$M parameters), MAX\_SEQ\_LEN${}=2048$, vocabulary size 8192.

\textbf{Dataset.} ClimbMix-400B, prepared with best-fit packing and BPE tokenization.

\textbf{Platforms.} (1) NVIDIA RTX 2060 (6\,GB VRAM) with PyTorch and \texttt{torch.compile}. (2) MacBook Pro with 96\,GB unified memory, Apple Silicon, MLX framework.

\textbf{Metric.} Validation bits per byte (val\_bpb), vocabulary-size-independent and architecture-independent (lower is better). All experiments use the same evaluation shard and token count.

\textbf{Time budget.} Fixed 300-second training budget per run. Curriculum overhead (difficulty scoring, buffer sorting, refill) counts against this budget.

\subsection{Experiment 1: Learning Rate Schedule Modulation}

We apply the King Wen sequence's surprise profile as a multiplicative modulation on the learning rate schedule. The 63 inter-hexagram surprise values are cycled across training steps, scaling the base learning rate by $(1 + A \cdot s_i)$ where $A$ is the amplitude and $s_i \in [-1, 1]$ is the centered surprise value. We test three amplitudes (0.15, 0.3, 0.5) against two controls: random perturbation (pseudo-random values at the same amplitude) and Shao Yong ordering (highly autocorrelated sawtooth pattern).

All runs use identical configuration: 4-layer GPT, standard warmup/warmdown LR envelope, RTX 2060, seed 42.

\begin{table}[h]
\centering
\begin{tabular}{lcc}
\toprule
\textbf{Schedule} & \textbf{val\_bpb} & \textbf{vs.\ Baseline} \\
\midrule
Baseline (no modulation) & 1.753 & --- \\
Random perturbation (amp${}=0.3$) & 1.731 & $-0.023$ \\
Shao Yong (amp${}=0.3$) & 1.732 & $-0.021$ \\
King Wen (amp${}=0.15$) & 1.777 & $+0.024$ \\
King Wen (amp${}=0.3$) & 1.785 & $+0.032$ \\
King Wen (amp${}=0.5$) & 1.790 & $+0.036$ \\
\bottomrule
\end{tabular}
\caption{LR schedule modulation results. King Wen degrades performance at all amplitudes.}
\label{tab:lr}
\end{table}

\textbf{Finding.} King Wen LR modulation degrades performance at all amplitudes tested (Table~\ref{tab:lr}). The degradation increases monotonically with amplitude. Both control perturbations show marginal improvements at the same amplitude, though as Section~\ref{sec:seeds} shows, these fall within natural seed variance.

\subsection{Experiment 2: Curriculum Ordering}

We reframe King Wen as a data ordering strategy rather than an LR modulator. Training batches are buffered (64 micro-batches), scored by a difficulty metric, and reordered. For the King Wen ordering, batches are mapped by rank: the batch at difficulty rank $k$ is placed at the King Wen position with surprise rank $k$.

Two difficulty metrics were used: token diversity (ratio of unique tokens to total tokens per batch) for initial CUDA experiments, and compression ratio (gzip compressed size / original size) for MLX and CUDA replication experiments.

\subsubsection{CUDA Results (RTX 2060, DEPTH=4, Token Diversity)}

\begin{table}[h]
\centering
\begin{tabular}{lcc}
\toprule
\textbf{Ordering} & \textbf{val\_bpb} & \textbf{vs.\ Sequential} \\
\midrule
Sequential (no buffer) & 1.719 & --- \\
Buffered passthrough & 1.680 & $-0.039$ \\
\textbf{Random shuffle} & \textbf{1.614} & $\mathbf{-0.106}$ \\
Easy-to-hard & 1.632 & $-0.087$ \\
Hard-to-easy & 1.627 & $-0.092$ \\
Shao Yong & 1.638 & $-0.081$ \\
King Wen & 1.662 & $-0.057$ \\
\bottomrule
\end{tabular}
\caption{CUDA curriculum ordering results. Random shuffle wins; King Wen is worst non-sequential.}
\label{tab:cuda}
\end{table}

All reorderings beat sequential (Table~\ref{tab:cuda}), but random shuffle wins. King Wen is the worst non-sequential ordering. The buffered passthrough control (buffering without reordering) itself improves performance by 0.039\,bpb, suggesting that the dataloader's best-fit packing creates sequential correlation that any disruption helps break.

\subsubsection{MLX Results (Apple Silicon, DEPTH=4, Compression Ratio)}

We repeat the experiment on MLX with five orderings (Shao Yong dropped as structurally similar to easy-to-hard). Two LR regimes are tested: standard warmdown and constant LR, following the finding that LR decay can suppress curriculum benefits \cite{luo2025lrdecay}.

\begin{table}[h]
\centering
\begin{tabular}{lcc}
\toprule
\textbf{Ordering} & \textbf{Standard WD} & \textbf{Constant LR} \\
\midrule
Sequential & 1.732 & 1.722 \\
Random & 1.713 & 1.697 \\
Easy-to-hard & 1.695 & 1.731 \\
Hard-to-easy & 1.709 & 1.707 \\
King Wen & 1.724 & 1.729 \\
\bottomrule
\end{tabular}
\caption{MLX DEPTH${}=4$ curriculum results. Noise floor: $\pm 0.060$\,bpb. No significant differences.}
\label{tab:mlx4}
\end{table}

Noise floor (3-seed range at DEPTH${}=4$): $\pm 0.060$\,bpb. \textbf{No ordering achieves a statistically significant improvement over sequential on MLX} (Table~\ref{tab:mlx4}). All differences fall within the measured noise floor.

\subsubsection{MLX Results (Apple Silicon, DEPTH=6)}

\begin{table}[h]
\centering
\begin{tabular}{lcc}
\toprule
\textbf{Ordering} & \textbf{Standard WD} & \textbf{Constant LR} \\
\midrule
Sequential & 2.056 & 2.039 \\
Random & 2.047 & 2.056 \\
Easy-to-hard & 2.099 & 2.079 \\
Hard-to-easy & 2.084 & 2.095 \\
King Wen & 2.074 & 2.030 \\
\bottomrule
\end{tabular}
\caption{MLX DEPTH${}=6$ curriculum results. Noise floor: $\pm 0.043$\,bpb.}
\label{tab:mlx6}
\end{table}

Noise floor at DEPTH${}=6$: $\pm 0.043$. King Wen achieves 2.030 under constant LR (Table~\ref{tab:mlx6}), a delta of $-0.044$, marginally outside noise, but this is a single run and does not replicate the CUDA pattern.

\subsubsection{Cross-Platform Analysis}

The large curriculum effect observed on CUDA (${\sim}0.04$--$0.11$\,bpb) does not replicate on MLX. The most likely explanation is that \texttt{torch.compile} creates kernel-level optimization patterns that overfit to sequential data ordering. Buffering disrupts this optimization, and random shuffling maximizes the disruption. This is a framework-specific artifact, not a property of the learning dynamics. The decorrelation benefit (breaking best-fit packing correlation) is real (${\sim}0.038$\,bpb from buffered passthrough) but is not King Wen-specific.

\subsection{Experiment 3: Seed Sensitivity Analysis}\label{sec:seeds}

To establish whether any of the above results exceed natural variance, we train 30 models with different random seeds (0--29) on the identical baseline configuration (4-layer GPT, standard LR schedule, RTX 2060, 5-minute budget).

\begin{table}[h]
\centering
\begin{tabular}{lc}
\toprule
\textbf{Statistic} & \textbf{Value} \\
\midrule
Range & 1.732--1.773 \\
Mean & 1.756 \\
Std & 0.009 \\
CV & 0.51\% \\
\bottomrule
\end{tabular}
\caption{val\_bpb across 30 random seeds. Natural variance range: 0.041\,bpb.}
\label{tab:seeds}
\end{table}

The seed sweep (Table~\ref{tab:seeds}) reveals that the ``improvements'' from random perturbation ($-0.023$) and Shao Yong ($-0.021$) in Experiment 1 fall within the natural seed variance range of 0.041\,bpb. Only King Wen's degradation ($+0.032$) is statistically meaningful---it exceeds the upper bound of observed seed variation.

Behavioral analysis of generated text across all 30 seeds (150 samples per seed, 4,500 total) shows that between-seed variance ratios are below 0.21 on all 12 metrics measured. PCA reveals a single ``verbosity axis'' (PC1 $= 66.4\%$ of between-seed variance), not multi-dimensional behavioral traits. Seeds produce quantitatively similar models at this scale.

\subsection{Discussion: Why Anti-Habituation Does Not Help}

The King Wen sequence's statistically significant properties---high transition distance, negative autocorrelation, full-spectrum variance---are precisely what make it harmful for gradient-based optimization:

\textbf{Excessive variance destabilizes gradient updates.} The sequence's defining statistical feature is higher variance than random permutations (Section~\ref{sec:stats}). When applied as LR modulation, this means larger effective perturbation magnitude than random noise at the same nominal amplitude.

\textbf{Negative autocorrelation prevents optimization momentum.} Shao Yong's highly autocorrelated pattern allows the optimizer to make consistent progress at a stable effective learning rate for several steps. King Wen is constructed to never repeat---each step reverses the previous step's intensity. This actively disrupts the temporal coherence that optimizers rely on.

\textbf{Anti-habituation is premature at small scale.} A 4-layer model trained for 5 minutes is in rapid early learning, far from the convergence plateau where habituation-breaking interventions might help. Disrupting a learner that hasn't plateaued simply slows it down.

\textbf{Fixed sequences cannot adapt.} Effective curriculum learning requires adaptation to learner state \cite{graves2017automated}. The King Wen sequence is a 3,000-year-old fixed ordering that cannot respond to the model's current loss landscape. Simple adaptive approaches---even random shuffling---outperform any fixed ordering because they do not impose incorrect assumptions about the learning trajectory.

\textbf{Framework artifacts confound curriculum evaluation.} The large CUDA curriculum effect (${\sim}0.11$\,bpb for random shuffle) disappears on MLX, revealing that \texttt{torch.compile}'s kernel optimization interacts with data ordering in ways that masquerade as learning dynamics. This is an important methodological finding: curriculum ordering experiments must be validated across frameworks, not just seeds.

%% ============================================================
\section{Implications}\label{sec:implications}
%% ============================================================

\textbf{For I-Ching studies.} The King Wen sequence has genuine combinatorial structure confirmed by Monte Carlo analysis. The four statistically significant properties---especially the yang-balanced groups ($p = 0.002$) and negative autocorrelation ($p = 0.037$)---are mathematical findings about a historically important object, independent of any ML application.

\textbf{For curriculum learning.} Fixed ordering curricula, even those with theoretically appealing statistical properties, do not substitute for adaptive curricula at the scales tested. The gap between combinatorial elegance and optimization utility is wide.

\textbf{For experimental methodology.} The \texttt{torch.compile} confound demonstrates that curriculum ordering experiments must be validated across frameworks. The 30-seed sweep (Section~\ref{sec:seeds}) provides a reusable template for distinguishing genuine effects from seed noise.

\textbf{For reproducibility.} All experiments use a fixed 5-minute time budget with val\_bpb as the single metric, making results directly comparable. The autoresearch framework's edit-train-evaluate loop provides a lightweight, reproducible experimental pipeline for testing training hypotheses.

%% ============================================================
\section{Limitations}\label{sec:limitations}
%% ============================================================

Our experiments are conducted at small scale: ${\sim}11.5$M parameter models trained for 5 minutes on a single GPU. Curriculum learning effects in the literature are typically demonstrated at 1B+ parameter scale with longer training. It is possible that King Wen's anti-habituation properties could help at larger scale, where models are more likely to reach convergence plateaus where habituation-breaking is relevant. However, the LR modulation results (Section~\ref{sec:experiments}) show degradation that increases with amplitude, not a marginal effect that might reverse at scale. The mechanistic explanation---excessive variance and negative autocorrelation disrupting gradient coherence---applies at any scale.

The \texttt{torch.compile} confound means our CUDA curriculum results reflect framework behavior, not pure learning dynamics. We address this by reporting both platforms, but the MLX null result means we lack a platform where curriculum ordering shows large, genuine effects against which to test King Wen.

%% ============================================================
\section{Conclusion}\label{sec:conclusion}
%% ============================================================

The King Wen sequence has statistically significant combinatorial properties. Monte Carlo analysis against 100,000 random permutations confirms four distinctive features: high transition distance, negative lag-1 autocorrelation, local yang balance, and pair-level asymmetry. These are not artifacts---they reflect genuine structure in a 3,000-year-old ordering of 64 binary states.

These properties do not translate to neural network training benefit. Learning rate modulation with the King Wen surprise profile degrades performance at all tested amplitudes. Curriculum ordering using the sequence is the worst non-sequential ordering on one platform and indistinguishable from noise on another. A 30-seed sweep confirms that only King Wen's degradation exceeds natural variance.

The central lesson is the gap between ``statistically interesting'' and ``useful for optimization.'' The King Wen sequence's high variance and negative autocorrelation---the properties that make it combinatorially distinctive---are counterproductive for gradient-based learning, which benefits from temporal coherence and moderate perturbation. Anti-habituation in a fixed sequence is not the same as adaptive curriculum design. We publish these negative results to close off a speculative research direction with data and to provide a methodological template for testing similar hypotheses.

The code, analysis tools, and full statistical results are available in the project repository.\footnote{\url{https://github.com/augchan42/king-wen-agi-framework}}

%% ============================================================
\bibliographystyle{plain}
\bibliography{references}

\newpage

%% ============================================================
\appendix
\section{Full Experimental Results}\label{app:results}
%% ============================================================

\subsection{LR Schedule Modulation (RTX 2060, DEPTH=4, Seed 42)}

\begin{table}[h]
\centering
\begin{tabular}{clccc}
\toprule
\textbf{Run} & \textbf{Schedule} & \textbf{Amplitude} & \textbf{val\_bpb} & \textbf{Memory (GB)} \\
\midrule
1 & Baseline & --- & 1.753 & 4.3 \\
2 & Random perturbation & 0.3 & 1.731 & 4.3 \\
3 & Shao Yong & 0.3 & 1.732 & 4.3 \\
4 & King Wen & 0.3 & 1.785 & 4.3 \\
5 & King Wen & 0.15 & 1.777 & 4.3 \\
6 & King Wen & 0.5 & 1.790 & 4.3 \\
\bottomrule
\end{tabular}
\end{table}

\subsection{Seed Sweep Summary (RTX 2060, DEPTH=4, 30 Seeds)}

Mean val\_bpb across 30 seeds: 1.756 (range 1.732--1.773, std 0.009, CV 0.51\%).

\subsection{Curriculum: CUDA with Compression Ratio (RTX 2060, DEPTH=4)}

\begin{table}[h]
\centering
\begin{tabular}{lcc}
\toprule
\textbf{Ordering} & \textbf{val\_bpb} & \textbf{vs.\ Sequential} \\
\midrule
Sequential & 1.778 & --- \\
Random & 1.627 & $-0.151$ \\
Easy-to-hard & 1.634 & $-0.144$ \\
Hard-to-easy & 1.634 & $-0.144$ \\
King Wen & 1.638 & $-0.140$ \\
\bottomrule
\end{tabular}
\end{table}

\subsection{Hardware Specifications}

\begin{table}[h]
\centering
\begin{tabular}{lllll}
\toprule
\textbf{Platform} & \textbf{GPU} & \textbf{VRAM} & \textbf{Framework} & \textbf{Precision} \\
\midrule
CUDA & NVIDIA RTX 2060 & 6\,GB & PyTorch + \texttt{torch.compile} & fp32 \\
MLX & Apple M-series & 96\,GB unified & MLX & fp32 \\
\bottomrule
\end{tabular}
\end{table}

%% ============================================================
\section{Code and Data Availability}\label{app:code}
%% ============================================================

Experimental code, ADR documents, and raw results are available at:

\begin{itemize}
\tightlist
\item King Wen AGI Framework: \url{https://github.com/augchan42/king-wen-agi-framework}
\item NVIDIA/CUDA experiments: documented in ADR-001 through ADR-008 in our autoresearch fork: \url{https://github.com/digital-rain-tech/autoresearch}
\item Apple Silicon (MLX) experiments: in our autoresearch-mlx fork: \url{https://github.com/digital-rain-tech/autoresearch-mlx}
\end{itemize}

The autoresearch framework by Andrej Karpathy provides the experimental infrastructure for the NVIDIA/CUDA experiments (our fork above); the Apple Silicon experiments use autoresearch-mlx, an MLX port by trevin-creator (\url{https://github.com/trevin-creator/autoresearch-mlx}) that we forked. All experiments use a fixed 5-minute training budget with val\_bpb as the single evaluation metric.

\end{document}